\documentclass[journal]{IEEEtran}

\usepackage[utf8]{inputenc} 
\usepackage[T1]{fontenc}    
\usepackage{hyperref}       
\usepackage{url}            
\usepackage{booktabs}       
\usepackage{amsfonts}       
\usepackage{nicefrac}       
\usepackage{microtype}      
\usepackage{lipsum}
\usepackage{graphicx}        
\usepackage{multicol}        
\usepackage{multirow}

\title{Sex-Classification from Cell-Phones Periocular Iris Images}

\author{
  Juan Tapia, Claudia Arellano, \\ Universidad Tecnologica de Chile - INACAP,\\ \texttt{j\_tapiaf@inacap.cl, clarellanov@inacap.cl} \\
   Ignacio Viedma, \\ Universidad Andres Bello - DCI,\\ \texttt{i.viedmaescalona@uandresbello.edu} \\
   ***Pre-print version accepted to be published On Selfie Biometrics Book-2019***.
}

\begin{document}
\maketitle

\begin{abstract}
Selfie soft biometrics has great potential for various applications ranging from marketing, security and online banking. However, it faces many challenges since there is limited control in data acquisition conditions. This chapter presents a Super-Resolution-Convolutional Neural Networks (SRCNNs) approach that increases the resolution of low quality periocular iris images  cropped from selfie images of subject's faces. This work shows that increasing image resolution  (2x and 3x) can improve the sex-classification rate when using a Random Forest classifier. The best sex-classification rate was 90.15\% for the right and 87.15\% for the left eye. This was achieved when images were upscaled from $150\times150$ to $450\times450$ pixels. These results compare well with the state of the art and show that when improving image resolution with the SRCNN the sex-classification rate increases. Additionally, a novel selfie database captured from 150 subjects with an iPhone X was created (available upon request).
\end{abstract}


\section{Introduction}
\label{sec:1}

Sex classification from images has become a hot topic for researchers in recent years since it can be applied to several fields such as security, marketing, demographic studies, among others. The most popular methods for sex classification are based on face, fingerprint and iris images. Iris based sex classification methods are usually based on Near Infra-Red (NIR) lighting and sensors. This has limited its use since it requires controlled environments and specific sensors.  Only recently has the literature explored the possibility of performing iris biometrics using color images \cite{ProencaAlexandre2007, Alonso-FernandezBigun2016,Sequeira14a, Nigam}. Color iris images are less suitable for classical iris processing algorithms since texture of dark-colored irides are not easily discernible in the VIS spectrum. 
In order to overcome this limitation, the inclusion of periocular information has been studied and shown to be one of the most distinctive regions of the face. This has allowed it to gain attention as an independent method for sex-classification or as a complement to face and iris modalities under non-ideal conditions. This region can be acquired largely relaxing the acquisition conditions, in contrast to the more carefully controlled conditions usually needed in NIR iris only systems. 

Results to date have not just shown the feasibility for sex-classification using VIS periocular iris images but have also reported the feasibility of acquiring other soft biometric information such as; for instance: ethnicity, age or emotion. \cite{Dantcheva2015}.

In this work, we proposed a method to classify sex from cell-phone (selfie) VIS periocular images. This is a challenging task since there is limited control of the quality of the images taken, since selfies can be captured from different distances, light conditions and resolutions (See. Figure 1). Cell-phones and Mobile devices in general have been widely used for communication, accessing social media, and also for sensitive tasks such as online banking. The use of soft biometrics such as sex classification in cell-phones may be useful for several applications. Real time electronic marketing, for instance,  may benefit from sex-classification by allowing web pages and Apps to offer products according to the person's sex. Data collection tasks may also benefit by discriminating target markets according to sex. Applications in security, on the other hand, may be highly improved by using sex-classification information. It may allow for the protection of users in tasks such as online banking, mobile payment and sensitive data protection. 

\begin{figure}[h]
\centering
\includegraphics[width=0.45\textwidth]{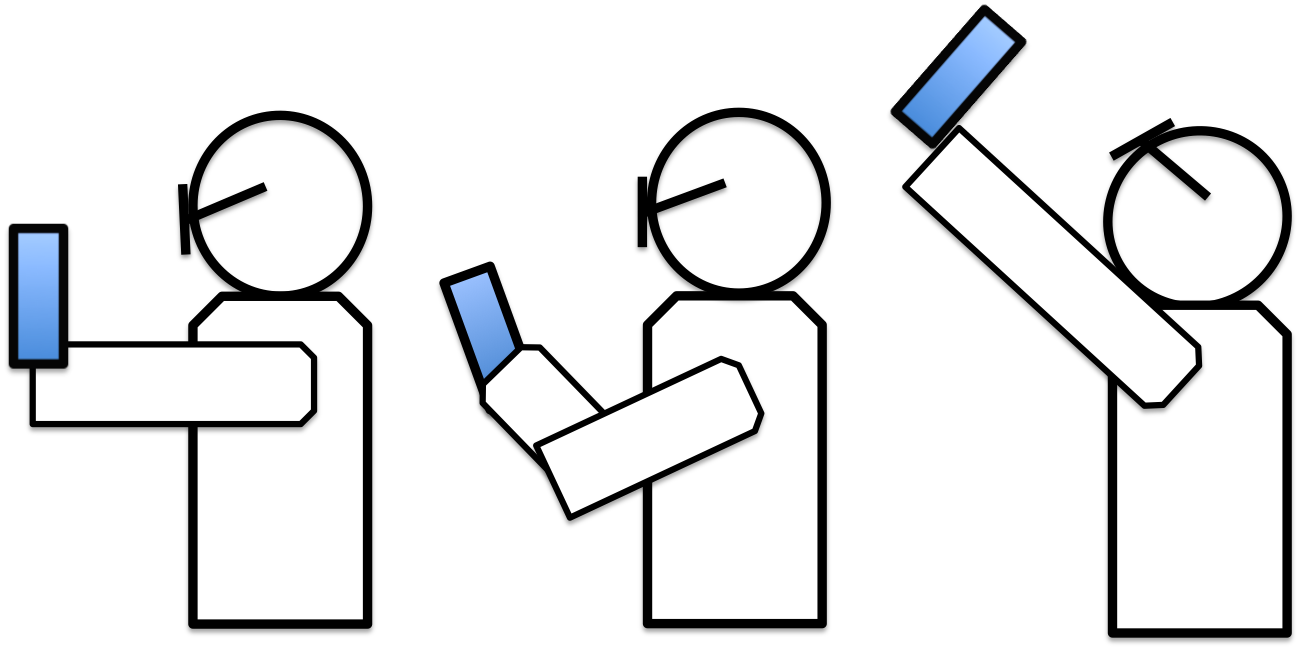}
\caption{ Representation of different conditions to capture the selfie images. Left: Straight arms. Middle: Half-Straight-arm. Right: Straight arm upper position.}
\label{fig:selfie}
\end{figure}

Previous work addressing biometric recognition on cell-phones include the use of additional accessories and products specially developed to facilitate this task. An example of such products is Aoptix Stratus\footnote{\url{http://www.ngtel-group.com/files/stratusmxds.pdf}}, a wrap around sleeve that facilitates NIR iris recognition on the iPhone. 

However, these products imply additional cost and only work for specific models of cell-phone (iphone). Therefore, it is important to study a reliable and user-friendly soft biometrics recognition system for all cellphone devices. Furthermore, as biometrics increasingly becomes more widely-used, the issue of interoperability is raised and the exchange of information between devices becomes an important topic of research to validate bio-metric results, since they should be indifferent to the sensor used to acquire the images \cite{Boyce2006, Pillai2014}.

\begin{figure*}
\centering
\includegraphics[width=0.8\textwidth]{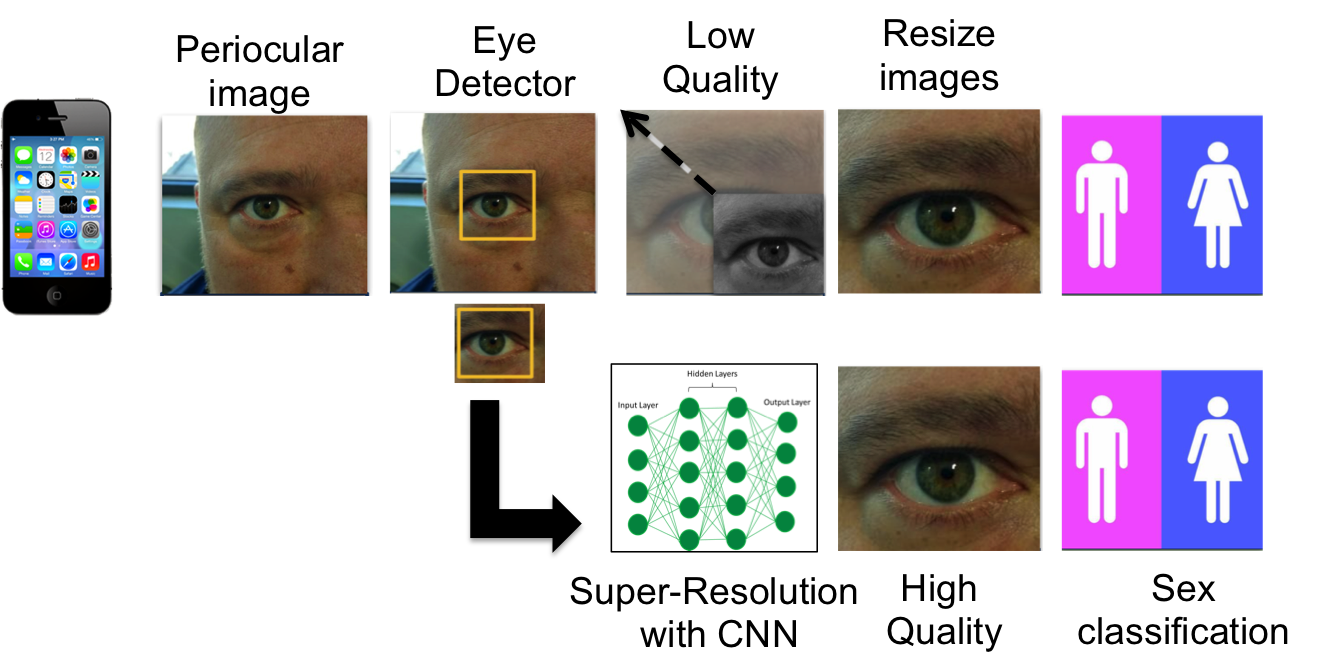}
\caption{ Block diagram of the proposed method. Top: The traditional sex classification approach. Botton: The proposed sex classification approach in order to improve the quality of the small images that comes from selfie images.}
\label{fig:Block_D}
\end{figure*}

Little work has been reported using periocular VIS cellphone images \cite{RattaniReddyDerakhshani2017}. They mainly use images that are cropped from selfies. In this context, the resulting periocular iris image has low quality resolution leading to weak sex-classification results. In this work,  a Convolutional Neural Network-Super-Resolution approach based on \cite{DongLoyHeEtAl2016} was proposed to limit this weakness as it allows the creation of a higher quality version of the same image (See Figure \ref{fig:Block_D}). The resulting  high resolution image is then used as input for a Random Forest algorithm that performs the sex-classification.

This approach is novel as there has not been previous attempts to classify sex from periocular iris cell-phone-images using super-resolution techniques for increasing the size of the low quality images that comes through selfies.

\subsection{Sex-classification from periocular VIS Images: State of the Art}\label{stateoftheart}

Sex classification from periocular VIS images has been reported multiple times in the literature \cite{AhujaIslamBarbhuiyaEtAl2016, TapiaViedma2017,TapiaAravena2017,Tapia2017}.   
Alonso-Fernandez et al. \cite{Alonso-FernandezBigun2016} reviewed the most commonly used techniques for sex-classification using periocular images. They also provided a comprehensive framework covering the most relevant issues in periocular images analysis. They presented algorithms for detecting and segmenting the periocular region, the existing databases, a comparison with face and iris modalities and the identification of the most distinctive regions of the periocular area among others topics. This work gives a comprehensive coverage of the existing literature on soft biometrics analysis from periocular images. A more recent review of periocular iris biometrics from the visible spectrum was made by Rattani et al. \cite{RattaniDerakhshani2017, RattaniDerakhshani2017a}. They addressed the subject in terms of computational image enhancement, feature extraction, classification schemes and designed hardware-based acquisition set-ups. 

Castrillon-Santana et al.\cite{Castrillon-SantanaLorenzo-NavarroRamon-Balmaseda2016} also proposed a sex-classification system that works for periocular images. They used a fusion of local descriptors to increase classification performance. They have also shown that the fusion of periocular and facial sex-classification reduces classification error. Experiments were performed on a large face database acquired in the wild where the periocular area was cropped from the face image after normalizing it with respect to scale and rotation.

Kumari et al. \cite{KumariBakshiMajhi2012} presented a novel approach for extracting global features from the periocular region of poor-quality grayscale images. In their approach, global sex features were extracted using independent component analysis and then evaluated using conventional neural-network techniques. All the experiments were performed on periocular images cropped from the FERET face database \cite{PhillipsWechslerHuangEtAl1998}. 

Tapia et al. \cite{TapiaAravena2018} trained a small convolutional neural network for both left and right eyes. They studied the effect of merging those models and compared the results against the model obtained by training a CNN over fused left-right eye images. They showed that the network benefits from this model merging approach, becoming more robust towards occlusion and low resolution degradation. This method outperforms the results obtained when using a single CNN model for the left and right set of images individually. 

Previous work addressing sex-classification is summarized in Table \ref{tab:table1}.

Several soft biometric approaches using periocular iris images captured from mobile devices such as cellphones are presented as follows. Zhang et al. \cite{ZhangLiSunEtAl2018} analyzed the quality of iris images on mobile devices. They showed that images are significantly degraded due to hardware limitations and the less constrained capture environment. The  identification rate using traditional algorithms is reduced when using these low-quality images. To enhance the performance of iris identification from mobile devices, they developed a deep feature fusion network that exploits complementary information from the iris and periocular regions. To promote iris recognition research on mobile devices under NIR illumination, they released the CASIA- Iris-Mobile-V1.0 database. 

Rattani et al \cite{RattaniReddyDerakhshani2017} proposed a convolutional neural network (CNN) architecture for the task of age classification. They evaluated the proposed CNN model on the ocular crops of the recent large-scale Adience benchmark for sex and age classification captured using smart-phones. The obtained results establish a baseline for deep learning approaches for age classification from ocular images captured by smart-phones. 

Raghavendra et al. \cite{RajaRaghavendraVemuriEtAl2015} demonstrated a new feature extraction method based on deep sparse filtering to obtain robust features for unconstrained iris images. To evaluate the proposed segmentation and feature extraction method, they employed an iris image database (VSSIRIS). This database was acquired using two different smartphones – iPhone 5S and Nokia Lumia 1020 under mixed illumination with unconstrained conditions in the visible spectrum. The biometric performance is benchmarked based on the equal error rate (EER) obtained from various state-of-art methods and a proposed feature extraction scheme.

\begin{table*}[h]
\scriptsize
\centering
\caption{Summary of sex-classification methods using images from eyes: I = Iris Images, P = Periocular Images, L = Left and R = Right, Acc= Accuracy.} \label{tab:table1}
\begin{tabular}{|c|c|c|c|c|c|c|}
\hline
Paper   & I/P    & Source & $N^o$ of Images &$N^o$ of Subjects& Type  & Acc (\%).  \tabularnewline
\hline
Thomas et al. \cite{Thomas2007}      & I       & Iris        & 16,469                    & N/A                        & NIR   & 75.00  \tabularnewline
\hline
Lagree et al. \cite{Lagree2011}          & I       & Iris        & 600                         & 300                       & NIR   & 62.17  \tabularnewline
\hline
Bansal et al. \cite{Bansal2012}          & I       & Iris       &400                         & 200                         & NIR  & 83.60  \tabularnewline
\hline
Juan E.Tapia et al. \cite{JuanE.Tapia2014}    & I       & Iris        &1,500                      & 1,500         &NIR   & 91.00 \tabularnewline
\hline
Costa-Abreu et al. \cite{Costa-Abreu2015}& I & Iris        &1,600                    & 200                         &NIR   & 89.74 \tabularnewline
\hline
Tapia et al. \cite{TapiaPerezBowyer2016} & I  & Iris        &3,000                    & 1,500                       &NIR   & 89.00  \tabularnewline
\hline
Bobeldyk et al. \cite{BobeldykRoss2016} & I / P  &  Iris   & 3,314 		& 1,083			  &NIR    & 85.70 (P)\\ & & & & &   &65.70 (I) \tabularnewline
\hline
Merkow et al. \cite{Merkow2010}	&P       &Faces		 & 936			  & 936			  & VIS   & 80.00 \tabularnewline
\hline
Chen et al. \cite{ChenRoss2011}	&P	   & Faces		 & 2,006				  &1,003		          & NIR/Thermal & 93.59 \tabularnewline
\hline
Castrillon et al. \cite{Castrillon-SantanaLorenzo-NavarroRamon-Balmaseda2016}& P  & Faces        &3,000                    & 1,500   &VIS & 92.46  \tabularnewline
\hline
Kuehlkamp et al. \cite{KuehlkampBeckerBowyer2017} & I                    & Iris                        & 3,000                                  & 1,500  & NIR  & 66.00 \tabularnewline
\hline
Rattani et al. \cite{RattaniReddyDerakhshani2017} & P &Faces & 572 & 200& VIS & 91.60\tabularnewline
\hline
Tapia et al. \cite{Tapia2017} & I                    & Iris                        & 10,000 unlabel &  --                           & NIR  & 77.79 \tabularnewline
&&&3,000 labeled&1,500& &83.00 \tabularnewline
\hline
Tapia et al. \cite{TapiaAravena2018} & P                    & Iris                        & 19,000  &  1,500                          & NIR  & 87.26 \tabularnewline

\hline
\end{tabular}
\end{table*}

\subsection{Challenges on VIS cell-phone periocular images}\label{VIS}

Selfie biometrics is a new topic only sparsely reported in the literature \cite{RattaniDerakhshani2017a}. 
Some of the aspects that make sex-classification from selfie images a challenging task are summarized as follows.
\subsubsection*{Cell-phone sensors }

The biometrics field is gradually becoming more and more part of daily life thanks to advances in sensor technology for capturing biometric data. More companies are producing and improving sensors for capturing periocular data \cite{Connaughton2012}. 

Most cameras are designed for RGB and their  quality can  suffer if they sense light in the IR part of the spectrum. IR blocking filters (commonly known as “hot mirrors”) are used in addition to Bayer patterns to remove any residual IR. This makes RGB sensors perform poorly when acquiring iris images. Specially when it comes to dark irises.

Due to space, power and heat dissipation limitations, camera sensors on mobile devices are much smaller than traditional iris sensors and the NIR light intensity is much weaker than that of traditional iris imaging devices. Therefore, the image noise on mobile devices is intensive, which reduces the sharpness and contrast of iris texture. 

Camera sensor size and focal length are small on mobile devices. As a result images of the iris are often less than 80 pixels in radius, which does not satisfy the requirement described in the international standard ISO/IEC 29794-6.2015 which restricts the iris pupil size to 120 pixels across iris diameters. 
Moreover, iris radius decreases rapidly as stand-off distance increases. The diameter of the iris decreases from 200 pixels to 135 pixels as the stand-off distance increases by only 10 cm. Although the iris radius in images captured at a distance is usually small, variation with distance is not so apparent because of long focal lengths.

\subsubsection*{Interoperability across sensors}
Several studies have investigated the interoperability of both face and fingerprint sensors. Additionally, there has been reports on sensor safety, illumination, and ease-of-use for iris recognition systems. As of writing, no studies have been conducted to investigate the interoperability of cell-phones cameras from various manufacturers using periocular information for sex classification algorithms. In order to function as a valid sex-classification system, texture sex patterns must prevail independent of the hardware used. The issue of interoperability among cell-phones is an important topic in large-scale and long-term applications of iris biometric systems \cite{Boyce2006, Connaughton2012, Pillai2014}.

\subsubsection*{Non controlled acquisition environment}

In non-constrained image capture settings such as the Selfie, it is not always possible to capture iris images with enough quality for reliable recognition under visible light. Periocular iris imaging from cross-sensors allows backward compatibility with existing databases and devices to be maintained while at the same time meet the demand for robust sex classification capability.
The use of the full periocular image helps overcome the limitations of just using iris information, improving classification rates \cite{KumariBakshiMajhi2012, Sequeira14a, Nigam}.  

Periocular cellphone images for biometrics applications are mainly coming from selfie face images. Traditionally, people capture selfie images in multiple places and backgrounds, using selfie sticks, alone or with others.  This translates to a high variability of images, in terms of size, light conditions and face pose in the image. To classify sex from a selfie, the periocular iris region from left, right or both eyes needs to be cropped. Therefore, resulting periocular images usually have very low resolution.

An additional limitation for cell-phones is size reduction when images are shared over the Internet. This may affect the accuracy of sex classification. For example, the Iphone X has a 7 MB selfie frontal camera. But images may be sent over the Internet using four size options: Small (60 kbytes), Medium (144 Kbytes), BIG(684 Kbytes) and real-size (2 MB).

In this work, a Super resolution CNN algorithm is proposed. This algorithm increases the resolution of images captured using cell-phones allowing better sex-classification rates. See Figure 2.

\begin{figure}[h]
\centering
\includegraphics[width=0.48\textwidth]{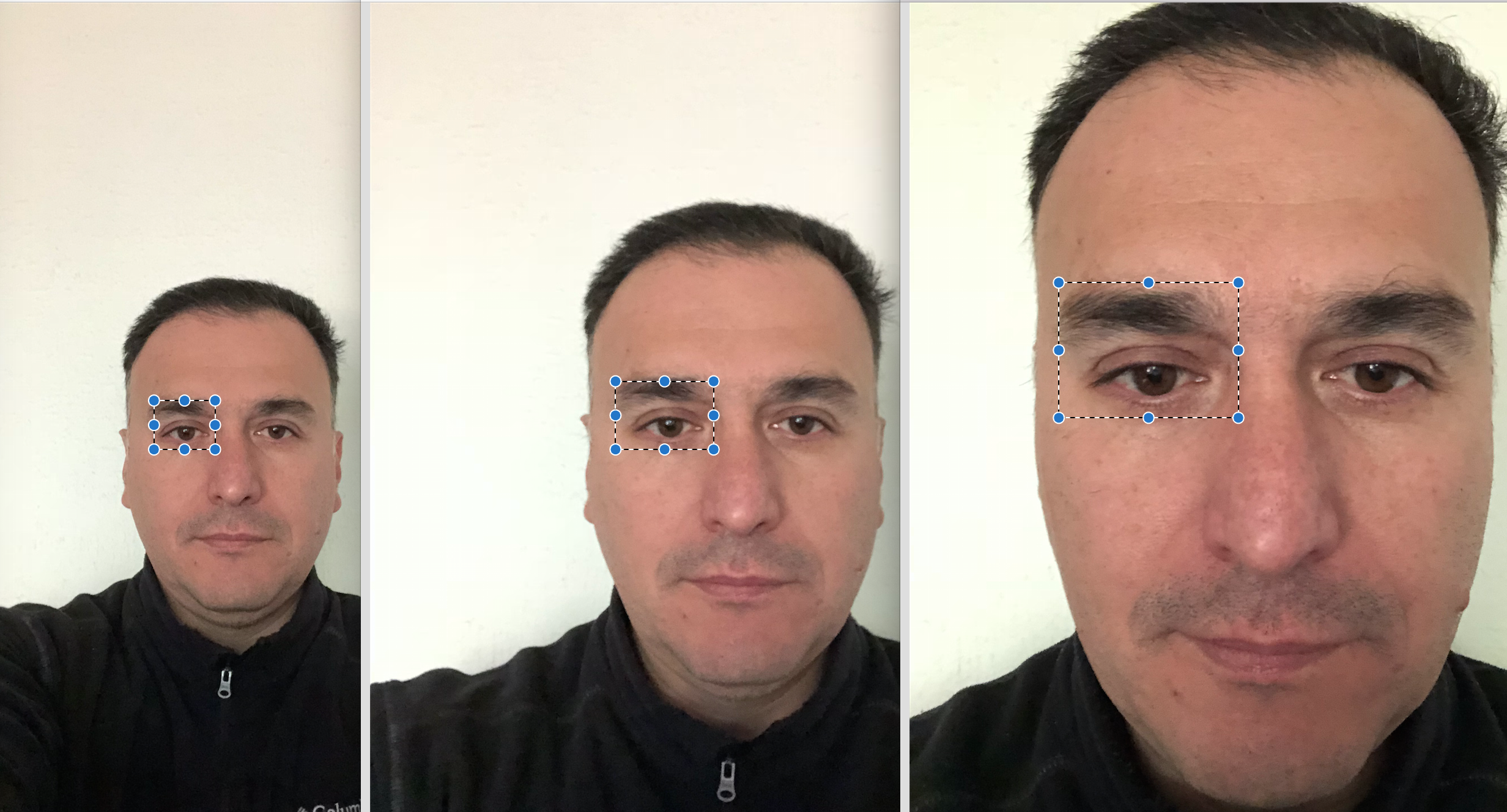}
\caption{ Example of selfie images captured from Iphone X with three different distances. Left: 1.0 mts (Straight arms). Middle: 60 cm (Middle straight arms). Right: 10 cm. (Arms close to the face). Dot squares show the periocular images. All images have the same resolution 2,320x3,088. }
\label{fig:lenet}
\end{figure}


\section {Proposed Method for Sex-Classification}
In this section a method for achieving sex-classification from cell-phone periocular images is described. The pipeline of this work is shown at the bottom of Figure \ref{fig:Block_D}. 
In Section \ref{SRCNN} the data Super Resolution Convolutional Neural Network algorithm used for resizing the images in order to increase their resolution is presented. The sex-classifier used  afterwards is a Random Forest algorithm which is describes in Section \ref{Classifier}.

\subsection{Super Resolution Convolutional Neural Networks}
\label{SRCNN}
Single-image super resolution algorithms can be categorized into four types: Prediction Models, Edge based Methods, Image Statistical Methods and Patch based (or example-based) Methods. These methods have been thoroughly investigated and evaluated in \cite{GlasnerBagonIrani2009, WeiXiaofengFangEtAl2018}. 

In this Chapter a patch based model to improve resolution of low quality images cropped from selfies is used. The Super-Resolution Using Deep Learning Convolutional Neural Networks (SRCNNs) algorithm proposed by Dong et al. \cite{DongLoyHeEtAl2016} was implemented. The network directly learns an end-to-end mapping between low and high-resolution images, with little pre/post processing beyond optimization.

\begin{figure}[h]
\centering
\includegraphics[width=0.48\textwidth]{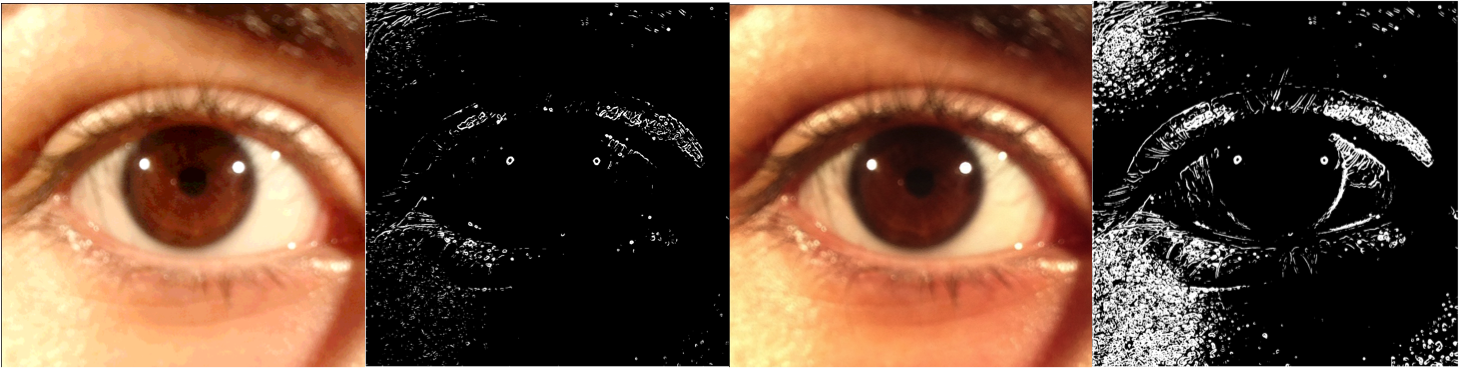}
\caption{ Example of feature maps of CONV1 and CONV2 layers. }
\label{fig:feature}
\end{figure}

The main advantage and most significant attributes of this method are as follows:

\begin{enumerate}
\item SRCNNs are fully convolutional, which is not to be confused with fully-connected. 
\item An image of any size (provided the width and height will tile) may be input into the algorithm making it very fast in comparison with traditional approaches.
\item It trains for filters, not for accuracy (See Figure \ref{fig:feature}).
\item They do not require solving an optimization problem on usage. After the SRCNN algorithm has learned a set of filters, a simple forward pass can be applied to obtain the super resolution output image. A loss function on a per-image basis does not have to be optimized to obtain the output. 
\item SRCNNs are entirely an end-to-end algorithm. The output is a higher resolution version of the input image. There are no intermediate steps. Once training is complete, the algorithm is ready to perform super resolution on any input image.
\end{enumerate}

The goal while implementing a SRCNNs algorithm is to learn a \textbf{set of filters} that allows low resolution inputs to be mapped to a higher resolution output. 
Two sets of image patches were created. One of them is a low resolution patch that is used as the input to the network. And  the second one a high resolution patch that will be the target for the network to predict/reconstruct. The SRCNN algorithm will learn how to reconstruct high resolution patches from low resolution input. Figure \ref{fig:feature} shows filter examples.

\subsection{Random Forest Classifier}
\label{Classifier}
To sex-classify (selfie) periocular images coming from different sensors (cell-phones) a Random Forest classifier (RF) was used. RF algorithm requires a single tuning parameter (Number of trees) making it simpler to use than SVM or Neural Network algorithms.  Furthermore, RF does not require a large amount of data for training like in Convolutional Neural Network algorithm.  

RF consists of a number of decision trees. Every node in the decision tree has a condition on a single feature and it is designed to split the dataset into two. The data with similar response values end up in the same set. 
The measure for the (locally) optimal condition is called impurity.
For classification, the most commonly used impurity measures are the Gini impurity (GDI), the Two Deviance Criterion (TDC) and the Twoing Rule (TR). The Gini's Diversity Index (GDI) can be expressed as follows:
 
 \begin{equation}
Gini\_index= 1-\sum_{i=1}=p^2(i)
\end{equation}

Where, the sum is over the classes $i$ at the node, and $p(i)$ is the observed fraction of classes with class $i$ that reach the node). A node with just one class (a pure node) has Gini index $0$; otherwise the Gini index is positive. 

The expression for the deviance of a node using the Two Deviance Criterion (TDC) is defined as follows:

 \begin{equation}
TDC\_index = -\sum_{i=1}=p(i) log p(i)
\end{equation} 
 
 The TR on the other hand, can be expressed as:  
 
\begin{equation}
 TR\_index = P(L)P(R)(\sum \mid L(i)-R(i) \mid) ^2
\end{equation} 

where $P(L)$ and $P(R)$ are the fractions of observations that split to the left and right of the tree respectively. If the result of the purity expression is large, the split make each child node purer. Similarly, if the expression is small, the split will make each child node more similar to each other, and hence similar to the parent node. Therefore, in this case the split does not increase the node purity. 

For regression trees, on the other hand, the impurity measure commonly used is the variance. When a tree is trained, the impact of each feature on the impurity of the node can be computed. This allows the features to be ranked according to the impurity measure.

\section{Experiments and Results}
\label{sec:4}
This section describes the experiments performed in order to evaluate sex classification from periocular VIS images. The databases used for the experiments are first introduced  in Section \ref{sec:3}. Additionally, a novel hand-made periocular iris image database captured from cellphones  (INACAP Database) is presented (available upon request).  Preprocessing and data-augmentation steps used for improving performance of the experiments are described in section \ref{DataAugmentation}. Section \ref{Hyper} describes  the process followed to determine the best parameters for the implementation of the SRCNN algorithm. Finally, in Section \ref{results} the experimental setup and results obtained are shown.

\subsection{Databases}
\label{sec:3}

One of the key problems for classifying soft-biometric features such as sex are the small quantity of sex-labeled images. Most databases available were collected for iris recognition applications. They do not, however, usually have soft-biometric information such as sex, age or ethnicity. In other cases, although this information may have been collected, it is not publicly available since it is considered private information. If only Selfie databases were considered, the lack of soft-biometric information is even worse. Most data available on the Internet are unlabeled images. The small amount of sex-labelled selfie images does not allow the training of powerful classifiers such as convolutional neural network and deep learning.

The existing databases used in this work and the novel INACAP database collected for this work are introduced as follows.

\subsubsection{Existing databases used for the experiments}

The following databases were used: CSIP \cite{SantosGranchoBernardoEtAl2015}, MICHE \cite{DeMarsicoNappiRiccioEtAl2015}, MODBIO \cite{Sequeira14a}. The \textbf{CSIP database} was acquired over cross-sensor setups and varying acquisition scenarios, mimicking the real conditions faced in mobile applications. It considered the heterogeneity of setups that cellphone sensor/lens can deliver (A total of 10 different setups). Four different devices (Sony Ericsson Xperia Arc S, iPhone 4,THL W200 and Huawei Ideos X3 (U8510)) were used and the images were captured at multiple sites. Where artificial, natural and mixed illumination conditions were used. Some of the images were captured using frontal/rear cameras and LED flash.

The \textbf{MICHE Database} captured images using smartphones and tablets such as the iPhone5 (IP5),Galaxy Samsung IV (GS4) and Galaxy Tablet II (GT2). 

The \textbf{MODBIO} database comprises the biometric data from 152 volunteers. Each person provided samples of face, iris and voice. There are 16 images for each person. The equipment used for  acquisition was a Portable hand held device, ASUS transformer Pad TF 300T, with the Android operating system. The device has two cameras one front and one back. The author used the back camera version TF300T-000128, with 8 MB resolution and autofocus. The sex distribution was 29\% females and 71\% males. Each image has a size of 640x480 pixels.

\subsubsection{Novel Home-made INACAP-database}

This database was collected by students from Universidad Andres Bello (UNAB) and Universidad Tecnologica de Chile - INACAP. This database contain 150 selfie images captured in three different distances according to the position from were the image was taken. We identify three possibles positions and classify the database accordingly:

\textbf{Set 1:} 150 selfies taken while the arm is extended up to front (Figure \ref{fig:selfie} Left)

\textbf{Set 2:} 150 selfies taken while the arm is bent towards the face (Figure \ref{fig:selfie} Middle)

\textbf{Set 3:} 150 selfies taken while the arm stretched up from the head (Figure \ref{fig:selfie} Right)

This is a person disjoint-dataset with 75 female and 75 male selfie images.
Table \ref{tab_vis} shows a summary of the databases used in this chapter.

\begin{table*}[]
\small
\centering
\caption{VW databases:  \emph{F} represents the number of Female images and \emph{M} the number of Male images; (*) only left images available.}\vspace{0.2cm}
\label{tab_vis}
\begin{tabular}{|c|c|c|c|l|l|c|}
\hline
\textbf{Dataset} & \multicolumn{1}{l|}{\textbf{Resolution}}                                                                                           & \multicolumn{1}{l|}{\textbf{No. Images}} & \multicolumn{1}{l|}{\textbf{No. Subjects}} & \textbf{F} & \textbf{M} & \textbf{Sensor(s)}                                                                              \\ \hline
CSIP(*) \cite{SantosGranchoBernardoEtAl2015}  & var. res. & 2,004 & 50                                     & 9          & 41         & \begin{tabular}[c]{@{}c@{}}Xperia ArcS, iPhone 4,\\ Th.I W200, Hua U8510\\ \end{tabular} \\ \hline
MOBBIO \cite{Sequeira14a}  & 250$\times$200  & 800 & 100 & 29  & 71  & Cell-phones \\ \hline
MICHE  \cite{DeMarsicoNappiRiccioEtAl2015}   & 1,000$\times$776 & 3,196 & 92   & 26 & 76 & iPhone 5 \\ \hline
Home-made   & 2,320$\times$3,088& 450& 150  & 75 & 75 & iPhone X \\ \hline
\end{tabular}
\end{table*}

\subsection{Data Preprocessing and Augmentation}
\label{DataAugmentation}

All the images from the databases used present different regions of interest as periocular images. OpenCV 2.10  was used to detect the periocular region and to normalize it by size. An eye detector algorithm was employed to automatically detect and crop the left and right periocular regions. All images were re-sized to 150x150 pixels. In those cases where the eye detector failed to select the periocular region, the image was discarded.

\begin{figure}[h]
\centering
\includegraphics[scale=0.40]{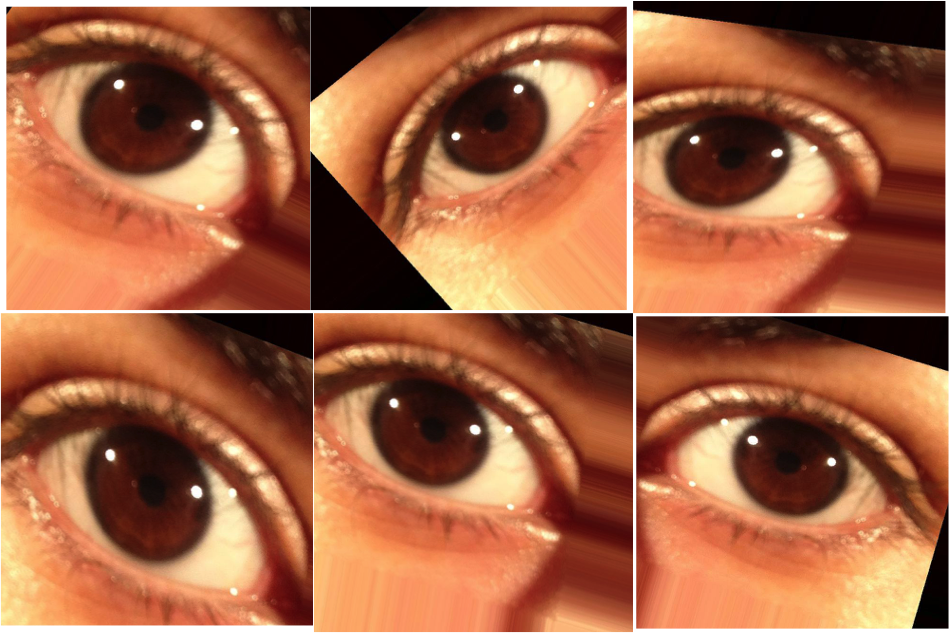}
\caption{ Data-augmentation examples used in order to increase the number of images available to train the classifier.}
\label{fig:da}
\end{figure}

To increase the number of images available from the left and the right eye an image generator function was used. The partition ratio for training, testing and validation sets was preserved. 
The dataset was increased from 6,000 to 18,000 images for each eye (36,000 images in total) using the following geometric transformations: Rotation (in ranges of 10 degrees), width and height Shifting (in ranges of 0.2) and Zoom Range of 15\%. All changes were made using the \textbf{Nearest fill mode}, meaning the images were taken from the corners to apply the transformation. The mirroring process was not applied since this \textbf{may transform} the left eye into a right eye. Care was taken not to mix training and testing examples. See Figure \ref{fig:da}.

\subsection{Hyper-parameters selection}
\label{Hyper}
A SRCNNs architecture that consists of only three CONV - RELU layers with no zero-padding was proposed. The first CONV layer learns 64 filters, each of which are $9 \times 9$. This volume is fed into a second CONV layer where $32$ filters of $1 \times 1$ were used to reduce dimensionality and learn local features. The final CONV layer learns a total of depth channels (which will be 3 for RGB images), each of which are $5 \times 5$. Finally, in order to measure the error rate, a mean-squared loss (MSE) rather than binary/categorical cross-entropy was used.

The rectifier activation function ReLU controls the non-linearity of individual neurons and when to activate them. There are several activation functions available. In this work, the suite of activation functions available on the Keras framework was evaluated. However, the best results for these CNNs were achieved where ReLU and Softmax activation functions were used.

In order to find the best implementation for the SRCNNs, the parameters of the CNN such as batch size, epoch, learning rate, among others needs to be determined.  

\paragraph{\textbf{Batch size:} Convolutional Neural Networks are in general sensitive to batch size, which is  the number of patterns shown to the network before the weights are updated. The batch size has an impact on training time and memory constraint. A set of different batch sizes from $n=16$ to $n=512$ in steps of $2^n$ were evaluated by the SRCNN algorithm.}

\paragraph{\textbf{Epochs:} The number of epochs is the number of times that the entire training dataset is shown to the network during training. The number of epochs were tested from $10$ up to $100$ in steps of $10$.}

\paragraph{\textbf{Learning Rate and momentum:} The Learning Rate (LR) controls how much the weights are updated at the end of each batch. The momentum, on the other hand,  controls how much the previous update is allowed to influence the current weight update. A small set of standard learning rates from the range $10e-1$ to $10e- 5$ and momentum values ranging from $0.1$ to $0.9$ in steps of $0.1$ were tried}

The selection of the best hyper-parameters of our modified implementation of SRCNN was found using a grid search fashion. The best classification rate was reached with a batch size of $16$, epoch number equal to: 50, LR of $1e-5$ and momentum equal to 0.9.

\begin{figure}[]
\centering
\includegraphics[width=0.48\textwidth]{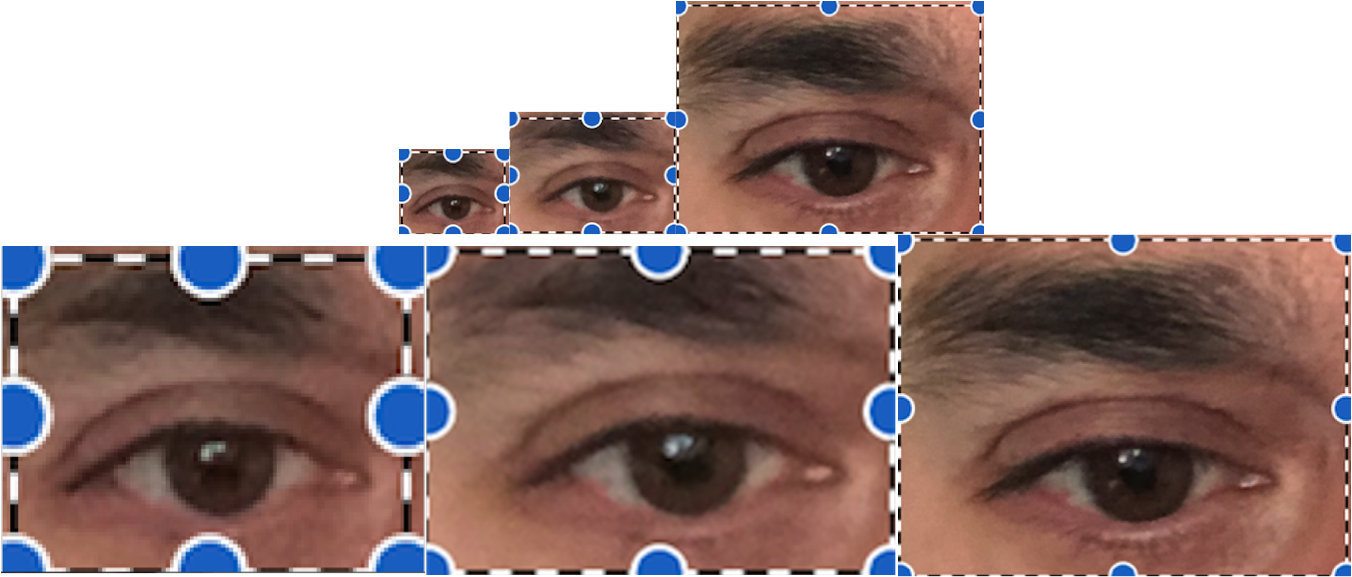}
\caption{ Top: Regular image cropped from face selfie image in three different scales and low quality images. Botton: Upscaling images generated from SRCNN in high quality images.}
\label{fig:srcnn1}
\end{figure}

According to the size of image used, a stride value equal to $15$ was proposed. The patch size used was $25\times25$ pixels. Figure \ref{fig:srcnn1} shows a example of input and output images from SRCNN.

\subsection{Experimental setup and results}
\label{results}

According to the pipeline shown in Figure \ref{fig:Block_D} there are two key processes involved to achieve sex-classification from periocular cellphone images. The Super resolution approach to increase resolution of images and the classifier itself.

For the super resolution process (SRCNNn) $3,000$ images taken from existing databases (CSIP, MICHE, MODBIO) were used as input. The algorithm generated $100,000$ patches of $25 \times 25$ pixels. This process allows the filters needed to achieve Super Resolution to be estimated. As result, the cropped selfie were transformed from its original dimension of $150\times150$ pixels to high resolution images of $300\times300$ (2X) and $450\times450$ (3X) pixels (See Figure \ref{fig:srcnn1}). 

The SRCNN algorithm was implemented using Keras and Theano (as  the  back end), both open source software libraries for  deep  learning. The training process was performed on an Intel i7 3.00 GHz processor and Nvidia P800 GPU. 

For the sex-classification process the Random Forest algorithm was used for all experiments using the three purity measure described in the previous section (Gini (GDI), Two Deviance Criterion (TDC) and the Twoing Rule (TR)). The algorithm was tested using several numbers from the tree (from 100 to 1000). For training, the databases was split into Left and Right eye images. For each eye, the existing databases were used (CSIP, MICHE, MODBIO ) with a total of 6,000 images plus the augmented data described in section \ref{DataAugmentation}. In total $18,000$ periocular images for each eye were used (Left and Right). For testing, the INACAP database which contains $450$ images was used.

Three experiments, were performed to evaluate the sex-classification rate. The first experiment (\textbf{Experiment 1}) was used as a baseline for comparison where the inputs are the original $150\times150$ pixel images. 

\textbf{Experiment 2} estimated the sex classification using the 2X up-scaled images from SRCNNs meaning the  $300\times300$ pixel images. 

\textbf{Experiment 3} used the 3X up-scaled images as input  ($450\times450$ pixel images).

\begin{table}[]
\centering
\caption{Sex classification results with Random Forest Classifier using a CSIP, MICHE and MODBIO dataset for trained and home-made dataset as validation dataset. SRCNN-X2 represents  of result with two times upscaling. SRCNN-X3 represents  of result with three times upscaling.} \label{tab:table_res}
\begin{tabular}{|l|l|l|l|l|l|l|l|}
\hline
\multirow{4}{*}{\textbf{Model}}   & \multirow{4}{*}{\textbf{Tree}} & \multicolumn{2}{c|}{\textbf{Traditional}}                                & \multicolumn{2}{c|}{\textbf{SRCNN-X2}}                                      & \multicolumn{2}{c|}{\textbf{SRCNN-X3}}                                      \\ \cline{3-8} 
                                  &                                & \multicolumn{2}{c|}{\textbf{150x150}}                                    & \multicolumn{2}{c|}{\textbf{300x300}}                                    & \multicolumn{2}{c|}{\textbf{450x450}}                                    \\ \cline{3-8} 
                                  &                                & \multicolumn{1}{c|}{\textbf{Left}} & \multicolumn{1}{c|}{\textbf{Right}} & \multicolumn{1}{c|}{\textbf{Left}} & \multicolumn{1}{c|}{\textbf{Right}} & \multicolumn{1}{c|}{\textbf{Left}} & \multicolumn{1}{c|}{\textbf{Right}} \\ \cline{3-8} 
                                  &                                & \multicolumn{1}{c|}{\textbf{(\%)}} & \multicolumn{1}{c|}{\textbf{(\%)}}  & \multicolumn{1}{c|}{\textbf{(\%)}} & \multicolumn{1}{c|}{\textbf{(\%)}}  & \multicolumn{1}{c|}{\textbf{(\%)}} & \multicolumn{1}{c|}{\textbf{(\%)}}  \\ \hline
\multirow{4}{*}{\textbf{RF-GDI}} & 100                            & 60.65                              & 62.60                               & 69.35                              & 72.15                               & 77.90                              & 78.90                               \\ \cline{2-8} 
                                  & 300                            & 61.35                              & 63.45                               & 68.70                              & 70.25                               & 78.90                              & 78.45                               \\ \cline{2-8} 
                                  & 500                            & 62.45                              & 64.45                               & 70.30                              & 73.40                               & 77.30                              & 79.15                               \\ \cline{2-8} 
                                  & 1,000                           & 66.70                              & 68.70                               & 74.45                              & 75.60                               & 79.90                              & 80.25                               \\ \hline
\multirow{4}{*}{\textbf{RF-TR}}   & 100                            & 62.25                              & 64.70                               & 75.20                              & 76.15                               & \textbf{83.45}                     & \textbf{83.45}                      \\ \cline{2-8} 
                                  & 300                            & 63.35                              & 66.50                               & 71.20                              & 75.80                               & \textbf{85.15}                     & \textbf{84.30}                      \\ \cline{2-8} 
                                  & 500                            & 64.45                              & 68.70                               & 74.45                              & 76.90                               & \textbf{86.30}                     & \textbf{88.90}                      \\ \cline{2-8} 
                                  & 1,000                           & 68.70                              & 71.00                               & 77.50                              & 77.15                               & \textbf{86.70}                     & \textbf{89.45}                      \\ \hline
\multirow{4}{*}{\textbf{RF-TDC}}  & 100                            & 63.35                              & 64.50                               & 74.00                              & 74.50                               & 78.90                              & 80.35                               \\ \cline{2-8} 
                                  & 300                            & 64.15                              & 65.90                               & 73.20                              & 75.90                               & 80.15                              & 84.15                               \\ \cline{2-8} 
                                  & 500                            & 64.56                              & 67.70                               & 75.50                              & 76.80                               & \textbf{84.05}                     & \textbf{89.25}                      \\ \cline{2-8} 
                                  & 1,000                            & 68.70                              & 70.15                               & 76.20                              & 76.50                               & \textbf{87.15}                     & \textbf{90.15}                      \\ \hline
\end{tabular}
\end{table}

The rate of sex classification obtained for all experiments is shown in Table \ref{tab:table_res}. Results for the Random Forest classifier using the three impurity measure and the following number of trees: 100, 300, 500 and 1,000 are also reported.
The best results for the  baseline experiment (Experiment 1) was 68.70\% and 71.00\% for the left and right periocular images respectively. Results improved as the image resolution increased. The best sex-classification rate (90.15\% for the right eye and 87.15\%) was achieved when 450x450 pixel images were used (SRCNN-3) and the RF algorithm was implemented using the TDC metric. These results are competitive with the state of the art and shows that when improving image resolution with SRCNN the sex-classification rate from periocular selfie images also improved.

\section{Conclusion}
\label{sec:5}
Selfie biometrics is a novel research topic that has great potential for multiple applications ranging from marketing, security and online banking. However, it faces numerous challenges to its use as there is only limited control over data acquisition conditions compared to traditional iris recognition systems, where the subjects are placed in specific poses in relation to the camera in order to capture an effective image. When using selfie images, we do not just deal with images taken from challenging environments, conditions and settings but also with low resolution since periocular image are mainly cropped from images of the entire face. 

This Chapter is preliminary work that demonstrates the feasibly of sex classification from cellphone (Selfie) periocular images. It has been shown that when using Super Resolution Convolutional Neural Networks for improving the resolution of periocular images taken from selfies, sex classification rates can be improved. 

In this work a Random Forest classifier algorithm was used. However, in order to move forward in this topic, it is necessary to create new sex-labelled databases of periocular selfie images. This would allow the use of better classifiers such as those based on deep learning. An additional contribution of this work, is a novel hand-made database (INACAP) that contains 450 sex-labeled selfie images captured with an iPhone X (Available upon request). 

\section{Acknowledgement}
\label{sec:6}
This research was funded by CONICYT, through grant FONDECYT - 1117089 And Universidad Tecnologica de Chile - INACAP. Further thanks to NVIDIA Research program to donate a P800 GPU card and make possible this research.



\bibliographystyle{unsrt}
\bibliography{References_ini_2}

\begin{thebibliography}{10}

\bibitem{ProencaAlexandre2007}
H.~Proenca and L.~A. Alexandre.
\newblock The nice.{I}: Noisy iris challenge evaluation - part {I}.
\newblock In {\em First IEEE International Conference on Biometrics: Theory,
  Applications, and Systems, 2007. BTAS 2007.}, pages 1--4, Sept 2007.

\bibitem{Alonso-FernandezBigun2016}
Fernando Alonso-Fernandez and Josef Bigun.
\newblock A survey on periocular biometrics research.
\newblock {\em Pattern Recognition Letters}, 82, Part 2:92 -- 105, 2016.
\newblock An insight on eye biometrics.

\bibitem{Sequeira14a}
Ana~F. Sequeira, João~C. Monteiro, Ana Rebelo, and Hélder~P. Oliveira.
\newblock {MobBIO: A Multimodal Database Captured with a Portable Handheld
  Device}.
\newblock In {\em Proc. of the 9th Int'l Conf. on Computer Vision Theory and
  Applications (VISIGRAPP 2014)}, pages 133--139, 2014.

\bibitem{Nigam}
Ishan Nigam, Mayank Vatsa, and Richa Singh.
\newblock Ocular biometrics.
\newblock {\em Inf. Fusion}, 26(C):1--35, November 2015.

\bibitem{Dantcheva2015}
{A}ntitza {D}antcheva, {P}etros {E}lia, and {A} {R}oss.
\newblock {W}hat else does your biometric data reveal? {A} survey on soft
  biometrics.
\newblock {\em {IEEE} {T}ransactions on {I}nformation {F}orensics and
  {S}ecurity, 2015, {ISSN}: 1556-6013}, 04 2015.

\bibitem{Boyce2006}
C.~Boyce, A.~Ross, M.~Monaco, L.~Hornak, and Xin Li.
\newblock Multispectral iris analysis: A preliminary study51.
\newblock In {\em Conference on Computer Vision and Pattern Recognition
  Workshop, 2006. CVPRW '06.}, pages 51--51, June 2006.

\bibitem{Pillai2014}
J.K. Pillai, M.~Puertas, and R.~Chellappa.
\newblock Cross-sensor iris recognition through kernel learning.
\newblock {\em IEEE Transactions on Pattern Analysis and Machine
  Intelligence,}, 36(1):73--85, Jan 2014.

\bibitem{RattaniReddyDerakhshani2017}
Ajita Rattani, Narsi Reddy, and Reza Derakhshani.
\newblock Convolutional neural network for age classification from smart-phone
  based ocular images.
\newblock In {\em 2017 {IEEE} International Joint Conference on Biometrics,
  {IJCB} 2017, Denver, CO, USA, October 1-4, 2017}, pages 756--761, 2017.

\bibitem{DongLoyHeEtAl2016}
C.~Dong, C.~C. Loy, K.~He, and X.~Tang.
\newblock Image super-resolution using deep convolutional networks.
\newblock {\em IEEE Transactions on Pattern Analysis and Machine Intelligence},
  38(2):295--307, Feb 2016.

\bibitem{AhujaIslamBarbhuiyaEtAl2016}
K.~Ahuja, R.~Islam, F.~A. Barbhuiya, and K.~Dey.
\newblock A preliminary study of cnns for iris and periocular verification in
  the visible spectrum.
\newblock In {\em 23rd International Conference on Pattern Recognition (ICPR)},
  pages 181--186, Dec 2016.

\bibitem{TapiaViedma2017}
J.~Tapia and I.~Viedma.
\newblock Gender classification from multispectral periocular images.
\newblock In {\em 2017 IEEE International Joint Conference on Biometrics
  (IJCB)}, pages 805--812, Oct 2017.

\bibitem{TapiaAravena2017}
Juan Tapia and Carlos Aravena.
\newblock Gender classification from nir iris images using deep learning.
\newblock In Bir Bhanu and Ajay Kumar, editors, {\em Deep Learning for
  Biometrics}, pages 219--239, Cham, 2017. Springer International Publishing.

\bibitem{Tapia2017}
Juan Tapia.
\newblock Gender classification from near infrared iris images, 2017.

\bibitem{RattaniDerakhshani2017}
Ajita Rattani and Reza Derakhshani.
\newblock Ocular biometrics in the visible spectrum: A survey.
\newblock {\em Image and Vision Computing}, 59:1 -- 16, 2017.

\bibitem{RattaniDerakhshani2017a}
Ajita Rattani and Reza Derakhshani.
\newblock On fine-tuning convolutional neural networks for smartphone based
  ocular recognition.
\newblock In {\em 2017 {IEEE} International Joint Conference on Biometrics,
  {IJCB} 2017, Denver, CO, USA, October 1-4, 2017}, pages 762--767, 2017.

\bibitem{Castrillon-SantanaLorenzo-NavarroRamon-Balmaseda2016}
Modesto Castrillon-Santana, Javier Lorenzo-Navarro, and Enrique
  Ramon-Balmaseda.
\newblock On using periocular biometric for gender classification in the wild:
  An insight on eye biometrics.
\newblock {\em Pattern Recognition Letters}, 82, Part 2:181 -- 189, 2016.

\bibitem{KumariBakshiMajhi2012}
Sunita Kumari, Sambit Bakshi, and Banshidhar Majhi.
\newblock Periocular gender classification using global \{ICA\} features for
  poor quality images.
\newblock {\em Procedia Engineering}, 38:945 -- 951, 2012.
\newblock International Conference On Modelling Optimization and Computing.

\bibitem{PhillipsWechslerHuangEtAl1998}
Phillips, Harry Wechsler, Jeffery Huang, and Patrick~J. Rauss.
\newblock The feret database and evaluation procedure for face-recognition
  algorithms.
\newblock {\em Image and Vision Computing}, 16(5):295--306, April 1998.

\bibitem{TapiaAravena2018}
J.~Tapia and C.~C. Aravena.
\newblock Gender classification from periocular nir images using fusion of cnns
  models.
\newblock In {\em 2018 IEEE 4th International Conference on Identity, Security,
  and Behavior Analysis (ISBA)}, pages 1--6, Jan 2018.

\bibitem{ZhangLiSunEtAl2018}
Q.~Zhang, H.~Li, Z.~Sun, and T.~Tan.
\newblock Deep feature fusion for iris and periocular biometrics on mobile
  devices.
\newblock {\em IEEE Transactions on Information Forensics and Security},
  13(11):2897--2912, Nov 2018.

\bibitem{RajaRaghavendraVemuriEtAl2015}
Kiran~B. Raja, R.~Raghavendra, Vinay~Krishna Vemuri, and Christoph Busch.
\newblock Smartphone based visible iris recognition using deep sparse
  filtering.
\newblock {\em Pattern Recognition Letters}, 57:33 -- 42, 2015.
\newblock Mobile Iris CHallenge Evaluation part I (MICHE I).

\bibitem{Thomas2007}
V.~Thomas, N.V. Chawla, K.W. Bowyer, and P.J. Flynn.
\newblock Learning to predict gender from iris images.
\newblock In {\em First IEEE International Conference on Biometrics: Theory,
  Applications, and Systems, BTAS 2007.}, pages 1--5, Sept 2007.

\bibitem{Lagree2011}
S.~Lagree and K.W. Bowyer.
\newblock Predicting ethnicity and gender from iris texture.
\newblock In {\em IEEE International Conference on Technologies for Homeland
  Security (HST),}, pages 440--445, Nov 2011.

\bibitem{Bansal2012}
A.~Bansal, R.~Agarwal, and R.~K. Sharma.
\newblock {SVM} based gender classification using iris images.
\newblock {\em Fourth International Conference on Computational Intelligence
  and Communication Networks (CICN),}, pages 425--429, Nov 2012.

\bibitem{JuanE.Tapia2014}
Juan~E. Tapia, Claudio~A. Perez, and Kevin~W. Bowyer.
\newblock Gender classification from iris images using fusion of uniform local
  binary patterns.
\newblock {\em European Conference on Computer Vision-ECCV, Soft Biometrics
  Workshop}, 2014.

\bibitem{Costa-Abreu2015}
M.~Da Costa-Abreu, M.~Fairhurst, and M.~Erbilek.
\newblock Exploring gender prediction from iris biometrics.
\newblock In {\em International Conference of the Biometrics Special Interest
  Group (BIOSIG), 2015}, pages 1--11, Sept 2015.

\bibitem{TapiaPerezBowyer2016}
J.~Tapia, C.~Perez, and K.~Bowyer.
\newblock Gender classification from the same iris code used for recognition.
\newblock {\em IEEE Transactions on Information Forensics and Security},
  PP(99):1--1, 2016.

\bibitem{BobeldykRoss2016}
Denton Bobeldyk and Arus Ross.
\newblock Iris or periocular? exploring sex prediction from near infrared
  ocular images.
\newblock In {\em Lectures Notes in Informatics (LNI), Gesellschaft fur
  Informatik, Bonn 2016.}, 2016.

\bibitem{Merkow2010}
J.~Merkow, B.~Jou, and M.~Savvides.
\newblock An exploration of gender identification using only the periocular
  region.
\newblock In {\em Fourth IEEE International Conference on Biometrics: Theory
  Applications and Systems (BTAS), 2010}, pages 1--5, Sept 2010.

\bibitem{ChenRoss2011}
Cunjian Chen and Arun Ross.
\newblock Evaluation of gender classification methods on thermal and
  near-infrared face images.
\newblock In {\em International Joint Conference on Biometrics (IJCB), 2011},
  pages 1--8. IEEE, 2011.

\bibitem{KuehlkampBeckerBowyer2017}
A.~Kuehlkamp, B.~Becker, and K.~Bowyer.
\newblock Gender-from-iris or gender-from-mascara?
\newblock In {\em 2017 IEEE Winter Conference on Applications of Computer
  Vision (WACV)}, pages 1151--1159, March 2017.

\bibitem{Connaughton2012}
R.~Connaughton, A.~Sgroi, K.~Bowyer, and P.J. Flynn.
\newblock A multialgorithm analysis of three iris biometric sensors.
\newblock {\em IEEE Transactions on Information Forensics and Security,},
  7(3):919--931, June 2012.

\bibitem{GlasnerBagonIrani2009}
D.~Glasner, S.~Bagon, and M.~Irani.
\newblock Super-resolution from a single image.
\newblock In {\em 2009 IEEE 12th International Conference on Computer Vision},
  pages 349--356, Sept 2009.

\bibitem{WeiXiaofengFangEtAl2018}
Z.~Wei, B.~Xiaofeng, H.~Fang, W.~Jun, and A.~A. Mongi.
\newblock Fast image super-resolution algorithm based on multi-resolution
  dictionary learning and sparse representation.
\newblock {\em Journal of Systems Engineering and Electronics}, 29(3):471--482,
  June 2018.

\bibitem{SantosGranchoBernardoEtAl2015}
Gil Santos, Emanuel Grancho, Marco~V. Bernardo, and Paulo~T. Fiadeiro.
\newblock Fusing iris and periocular information for cross-sensor recognition.
\newblock {\em Pattern Recogn. Lett.}, 57(C):52--59, May 2015.

\bibitem{DeMarsicoNappiRiccioEtAl2015}
Maria De~Marsico, Michele Nappi, Daniel Riccio, and Harry Wechsler.
\newblock Mobile iris challenge evaluation (miche)-i, biometric iris dataset
  and protocols.
\newblock {\em Pattern Recogn. Lett.}, 57(C):17--23, May 2015.

\end{thebibliography}
\end{document}